\pdfoutput=1
\documentclass[10pt,twocolumn,letterpaper]{article}

\usepackage{cvpr}
\usepackage{times}
\usepackage{epsfig}
\usepackage{graphicx}
\usepackage{amsmath}
\usepackage{amssymb}
\usepackage{times}
\usepackage{epsfig}
\usepackage{graphicx}
\usepackage{amsmath}
\usepackage{amssymb}
\usepackage[utf8]{inputenc} % allow utf-8 input
\usepackage[T1]{fontenc}    % use 8-bit T1 fonts
\usepackage{url}            % simple URL typesetting
\usepackage{booktabs}       % professional-quality tables
\usepackage{amsfonts}       % blackboard math symbols
\usepackage{nicefrac}       % compact symbols for 1/2, etc.
\usepackage{microtype}      % microtypography
\usepackage{times}
\usepackage{epsfig}
\usepackage{graphicx}
\usepackage{amsmath}
\usepackage{amssymb}
\usepackage{multirow}
\usepackage{booktabs} % Allows the use of \toprule, \midrule and
\usepackage{algorithm}
\usepackage{algorithmic}
\usepackage{subfigure}

\usepackage{dsfont}
\cvprfinalcopy % *** Uncomment this line for the final submission

\def\cvprPaperID{319} % *** Enter the CVPR Paper ID here

\ifcvprfinal\fi
\begin{document}

%%%%%%%%% TITLE
\title{Dynamic-structured Semantic Propagation Network%: A Universal Semantic Segmentation Model
}

\author{Xiaodan Liang\\
Carnegie Mellon University\\
{\tt\small xiaodan1@cs.cmu.edu}
\and
Hongfei Zhou\\
Wumii Tech. Limited\\
{\tt\small hfzhou@wumii.com}
\and
Eric Xing\\
Petuum Inc\\
{\tt\small eric.xing@petuum.com}
}

\maketitle
%\thispagestyle{empty}

%%%%%%%%% ABSTRACT
\begin{abstract}

Semantic concept hierarchy is still under-explored for semantic segmentation due to the inefficiency and complicated optimization of incorporating structural inference into dense prediction. This lack of modeling semantic correlations also makes prior works must tune highly-specified models for each task due to the label discrepancy across datasets. It severely limits the generalization capability of segmentation models for open set concept vocabulary and annotation utilization. In this paper, we propose a Dynamic-Structured Semantic Propagation Network (DSSPN) that builds a semantic neuron graph by explicitly incorporating the semantic concept hierarchy into network construction. Each neuron represents the instantiated module for recognizing a specific type of entity such as a super-class (e.g. food) or a specific concept (e.g. pizza). During training, DSSPN performs the dynamic-structured neuron computation graph by only activating a sub-graph of neurons for each image in a principled way. A dense semantic-enhanced neural block is proposed to propagate the learned knowledge of all ancestor neurons into each fine-grained child neuron for feature evolving. Another merit of such semantic explainable structure is the ability of learning a unified model concurrently on diverse datasets by selectively activating different neuron sub-graphs for each annotation at each step. Extensive experiments on four public semantic segmentation datasets (i.e. ADE20K, COCO-Stuff, Cityscape and Mapillary) demonstrate the superiority of our DSSPN over state-of-the-art segmentation models. Moreoever, we demonstrate a universal segmentation model that is jointly trained on diverse datasets can surpass the performance of the common fine-tuning scheme for exploiting multiple domain knowledge.
   
\end{abstract}

%%%%%%%%% BODY TEXT
\section{Introduction}

Recognizing and segmenting arbitrary objects, posed as a primary research direction in computer vision, has achieved great success driven by the advance of convolutional neural networks (CNN). However, current segmentation models using generic deeper and wider network layers~\cite{long2015fully,chen2016deeplab,zhao2016pyramid,wu2016wider,lin2016refinenet} still show unsatisfactory results of recognizing objects in a large concept vocabulary with limited segmentation annotations. The reason is that they ignore the intrinsic taxonomy and semantic hierarchy of all concepts. For example, \emph{giraffe}, \emph{zebra} and \emph{horse} categories share one super-class \emph{ungulate} that depicts their common visual characteristics, which makes them be easily distinguished from \emph{cat}/\emph{dog}. In addition, due to diverse level of expertise and application purposes, the target concept set of semantic segmentation can be inherently open-ended and highly structured for each specific task/dataset. However, some few techniques also explored the semantic hierarchy for visual recognition by resorting to complex graphical inference~\cite{deng2014large}, hierarchical loss~\cite{redmon2016yolo9000} or word embedding priors~\cite{zhao2017open} on final prediction scores. Their loss constraints can only indirectly guide visual features to be hierarchy-aware, which is hard to be guaranteed and often leads to inferior results compared to generic CNN models. 

Furthermore, this lack of modeling semantic hierarchy also prohibits the research towards a universal segmentation model that can address the segmentation of all concepts at once. Existing works~\cite{long2015fully,chen2016deeplab,zhao2016pyramid,wu2016wider} often strive to train a task-specific model due to the label discrepancy across dataset with limited annotations. That way largely limits the model generation capability and deviates from human perception that can recognize and associate all concepts by considering the concept hierarchy. If one wants to improve one task by fully utilizing other annotations with different label set, prior works must remove the classification layer and only share intermediate representations. Our target of learning a universal segmentation model also has some connections to very recent researches in combining different visual tasks~\cite{kokkinos2016ubernet,wang2016cnn} or multi-modal tasks~\cite{kaiser2017one} in one model, which often use several fixed network branches with specialized losses to integrate all tasks. 

\begin{figure*}[!tp]
		\begin{center}
			\includegraphics[scale=0.55]{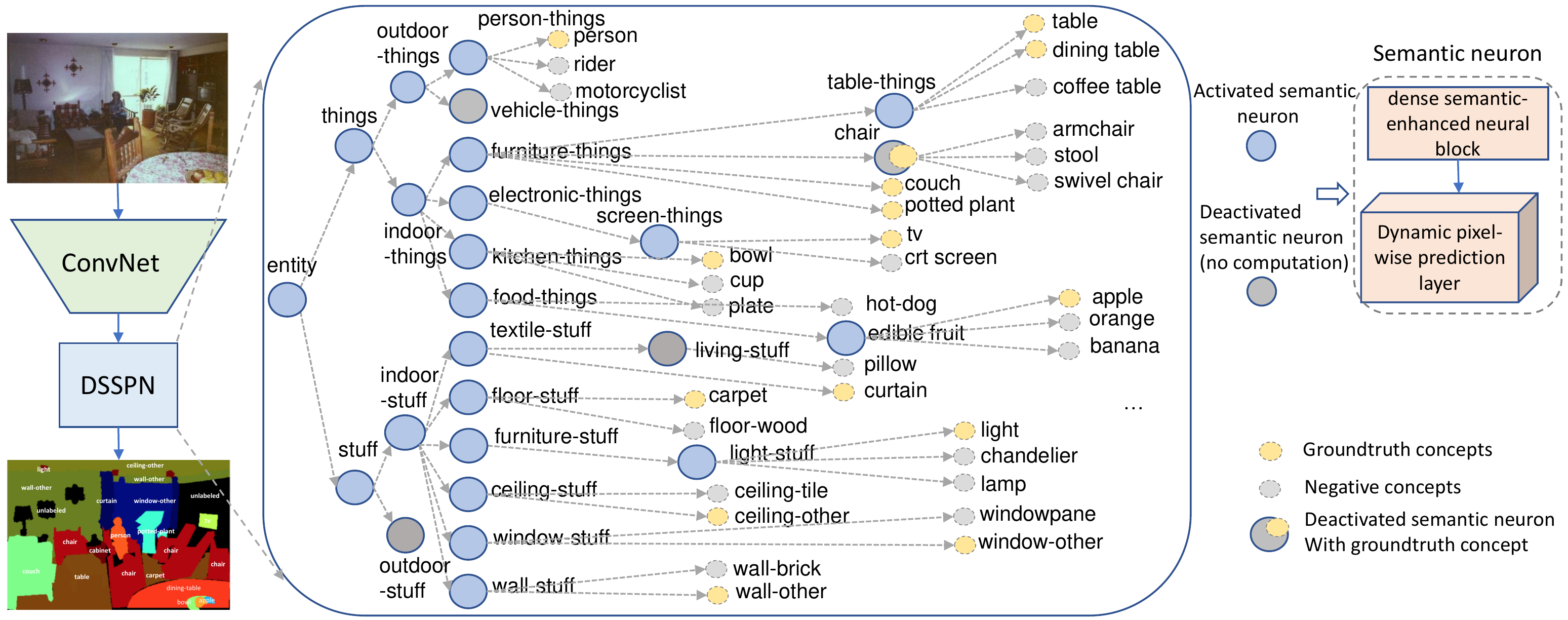}
			\caption{An overview of the proposed DSSPN that explicitly models dynamic network structures according to a semantic concept hierarchy. The basic convolutional features are propagated into a dynamic-structured semantic neuron graph for hierarchical pixel-wise recognition. During training, DSSPN only activates a sub-graph of semantic neurons that reach out the target labels for each image, leading to dynamic-structured feed-forward propagation and back-propagation. It means DSSPN only needs to focus on hierarchically classifying confusing concepts with the same parent during training. For clarity, we only show a portion of semantic neurons.}
			\label{fig:framework}
		\end{center}
		\vspace{-4mm}
	\end{figure*}
	
 In this work, we aim at explicitly integrating a semantic concept hierarchy into the dynamic network optimization, called as Dynamic-Structured Semantic Propagation Network (DSSPN). In the spirit of curriculum learning~\cite{bengio2009curriculum} that gradually increases the target difficulty levels and exploits previously learned knowledge for learning new fine-grained concepts, DSSPN first progressively builds a semantic neuron graph following the semantic concept hierarchy in which each neuron is responsible for segmenting out regions of one concept in the word hierarchy. The learned features of each neuron are further propagated into its child neurons for evolving features in order to recognize more fine-grained concepts. For each image or dataset, DSSPN performs the dynamic-structured semantic propagation over an activated semantic neuron sub-graph where only the present concepts and their ancestors are selected. Benefiting from the merits of semantically ordered network modules and the dynamic optimization strategy, our DSSPN would enable the learned visual representation to naturally embed rich semantic correlations between diverse concepts. Such explicit neuron definition mechanism makes the proposed DSSPN be a semantically explainable dynamic network architecture with good memory and computation efficiency. 
 
 Rather than only taking into account features of the parent neuron for each neuron, we introduce a new dense semantic-enhanced neural block which densely integrates the features of all ancestor neurons to evolve feature representation of each neuron, inspired by DenseNets~\cite{huang2016densely}. By broadcasting the learned knowledge of all ancestor neurons into each neuron, our DSSPN can fully exploit the semantic correlation and inheritance into the feature learning in a more efficient way. As explained in very recent information bottleneck theory~\cite{tishby2015deep}, the deep networks often tend to squeeze the information through a bottleneck and retain only the features most relevant to targets. Such dense semantic connection thus alleviates the information loss along deeper layers by explicitly enforcing ancestor neurons to preserve discriminate features for recognizing more fine-grained concepts. 
 
 Note that our DSSPN activates dynamic computation graphs for each sample during training. For scalability, a dynamic batching optimization scheme is proposed to enable optimize multiple computation graphs within one batch by configuring a dynamic number of samples for learning distinct neural modules at each step. A memory efficient implementation of our DSSPN is also described.
 
 Extensive experiments on four popular semantic segmentation datasets (i.e. Coco-Stuff~\cite{caesar2016coco}, ADE20k~\cite{zhou2016semantic}, Cityscape~\cite{cordts2015cityscapes} and Mapillary~\cite{neuhold2017mapillary}) demonstrate the effectiveness of incorporating our DSSPN into the state-of-the-art basic segmentation networks. We thus demonstrate that our dynamic-structure propagation mechanism is an effective way to implement a semantic explaining way that is needed for segmenting massive intrinsically structured concepts. Moreover, we show that learning a unified DSSPN model over diverse models can also bring the performance over the commonly used fine-tuned scheme for utilizing annotations in multiple domains.

\section{Related Work}

\textbf{Semantic segmentation.} Semantic segmentation has recently attracted a hug amount of interests and achieved great progress with the advance of deep convolutional neural networks. Most of prior works focus on developing new structures and filter designs to improve general feature representation, such as deconvolutional neural
network~\cite{noh2015learning}, encoder-decoder architecture~\cite{badrinarayanan2015segnet}, dilated convolution~\cite{chen2016deeplab,yu2015multi}, pyramid filters~\cite{zhao2016pyramid} and wider nets~\cite{wu2016wider}. Although these methods show promising results on datasets with a small label set, e.g. 21 of PASCAL VOC~\cite{everingham2010pascal}, they obtain relatively low performance on recently released benchmarks with large concept vocabularies (e.g. 150 of ADE20k~\cite{zhou2016semantic} and 182 of COCO-stuff~\cite{caesar2016coco}). These models directly use one flat prediction layer to classify all concepts and disregard their intrinsic semantic hierarchy and correlations. Such prediction strategy largely limited the model capability and also makes the network parameters hardly adapt to other recognition tasks or new objects. In this paper, our DSSPN builds the dynamic network structure according the semantic concept hierarchy, where each neural module takes care of recognizing one concept in the taxonomy, and modules are connected following the structure to enforce semantic feature propagation.

\textbf{Dynamic and graph network structure.} Exploring dynamic networks has recently received increasing attentions due to their good model flexibility and huge model capacity. Prior works proposed a family of graph-based CNNs~\cite{niepert2016learning}, RNNs~\cite{liang2016semantic} and reinforcement learning structures~\cite{liang2017deep} to accommodate networks into different graph-structured data, such as superpixels, social networks and object relationships. There exists some few works that investigated dynamic networks. For example, Liang et al.~\cite{liang2017interpretable} evolved the network structures by learning how to merge the graph nodes automatically. Shi et al.~\cite{shi2017deep} aims at learning the local correlation structure for spatio-temporal data. Different from them, our DSSPN introduces a general dynamic network for recognizing and segmenting out objects in the large-scale and highly-structured concept vocabulary. The neural modules are dynamically activated following the present concept tree for each image.

\textbf{Hierarchical recognition.} There is a line of researches that exploit the structure of WordNet to achieve hierarchical recognization. For example, Deng et al.~\cite{deng2012hedging} used an accuracy-specificity trade-off algorithm to explore the WordNet hierarchy while Ordonez et al.~\cite{ordonez2013large} learns the mapping of common concepts to entry-level concepts. Deng
et al.~\cite{deng2014large} further employed a label relation graph to guide the neural network learning. Most recently, Zhao et al.~\cite{zhao2017open} addressed the open-vocabulary scene parsing by constructing asymmetric word-embedding space. Rather than implicitly enforcing semantic relations into network representations as previous works did, the proposed DSSPN explicitly constructs the network modules guided by their semantic hierarchy. The dynamic neural activation strategy makes the model scalable and applicable for a universal segmentation model.

\begin{figure}[!tp]
		\begin{center}
			\includegraphics[scale=0.52]{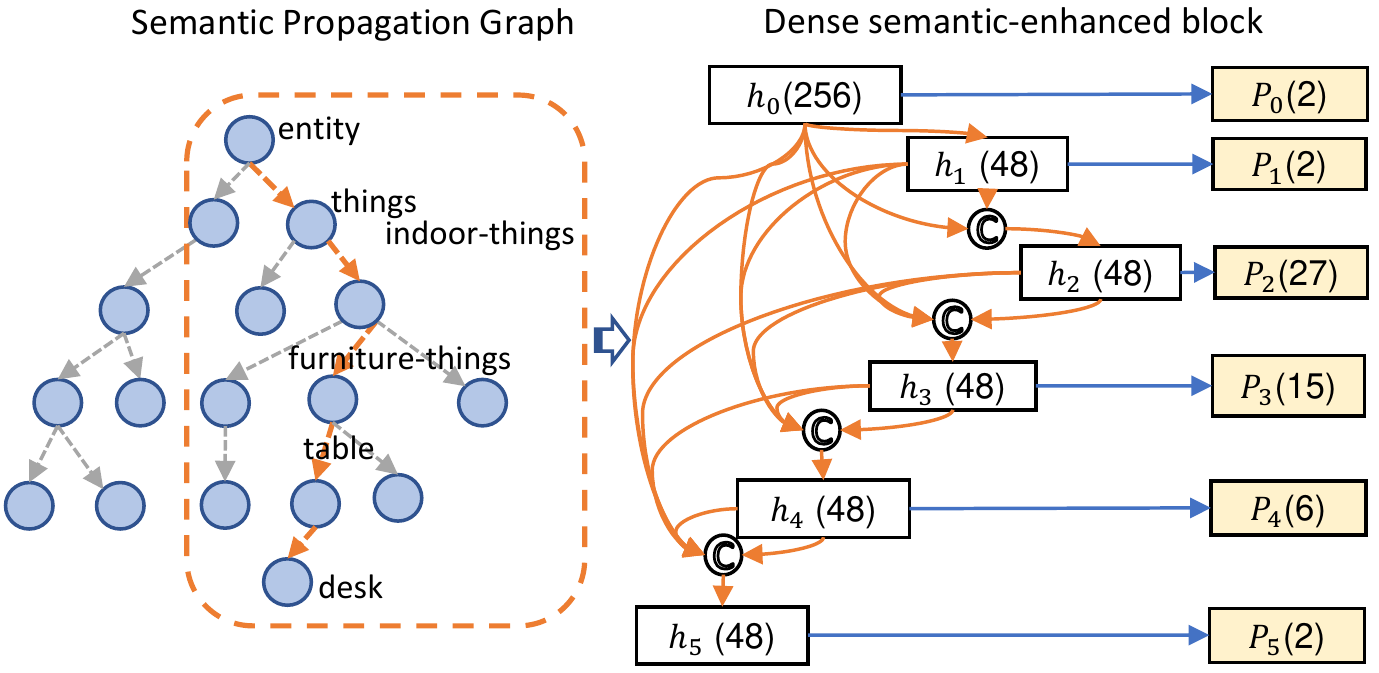}
			\caption{Dense semantic-enhanced block. For each activated semantic neuron $v_i$, it concatenates features evolved from the neurons in the whole path (dashed orange arrows) to obtain enhanced representation $h_i$, which is further passed through dynamic pixel-wise prediction $P_i$ for distinguishing between its children nodes. The output dimensions (e.g. 48 ) of each block and those (e.g. 27) of pixel-wise prediction layer are shown in the parentheses. }
			\label{fig:dynamic}
		\end{center}
		\vspace{-5mm}
	\end{figure}
	
\section{Dynamic Tree-Structured Propagation Network}
\label{sec:framework}
In order to fully exploit concept correlations for recognizing and segmenting out a large-scale concept vocabulary, we aim at explicitly incorporating the semantic concept hierarchy into the dynamic network structure for semantic segmentation. Figure~\ref{fig:framework} gives an overview of our proposed DSSPN. After feeding the images into basic convolutional networks for extracting intermediate features $x$, the DSSPN is appended to perform dense pixel-wise recognition with a dynamically induced neural propagation scheme. We first build a large semantic neuron graph that each neuron corresponds to one parent concept in the semantic concept hierarchy and aims at recognizing between its child concepts. During training, given the concepts appeared in each image, only a small neuron graph that would derive the target concepts are activated, leading to the dynamic semantic propagation graph for effective and efficient computation. A new dense semantic-enhanced neural block is proposed to evolve features for fine-grained concepts by incorporating features of their ancestor concepts.  We describe in more details in the following sections.

\subsection{Semantic Neuron Graph} 

\begin{figure*}[!tp]
		\begin{center}
			\includegraphics[scale=0.58]{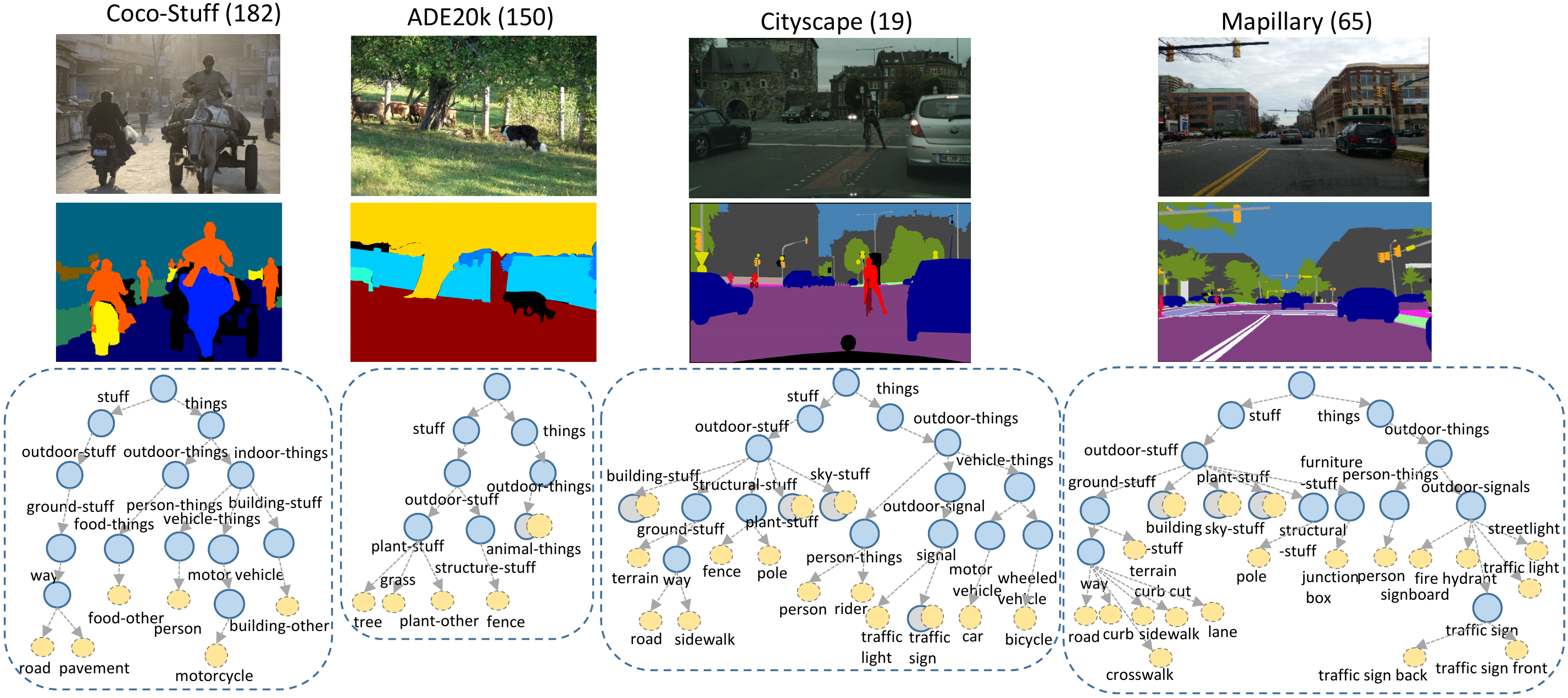}
			\caption{DSSPN can learn a unified segmentation model for accommodating diverse annotation policies. For training with diverse annotations with discrepant label granularity, DSSPN activates dynamic-structured semantic propagation graphs for each image. For example, Ade20k only annotates a single ``animal-thing" label for all animal categories while Cocostuff elaborately categories each fine-grained concept, e.g. cat or elephant. The semantic neurons that correspond to target labels are deactivated (grey colored solid circles). It thus fully exploits the shared concept patterns and concept relationship in a semantic hierarchy for a more general segmentation model. For simplicity, only target labels and their ancestor concepts are shown.}
			\label{fig:universal}
		\end{center}
		\vspace{-6mm}
	\end{figure*}
	
We first denote the semantic concept graph as $\mathcal{G}^c = <\mathcal{C}^v, \mathcal{C}^e>$, where $\mathcal{C}^v$ consists of all concepts $\{\mathcal{C}^v_i\}$ in a predefined knowledge graph (described in Section~\ref{sec:exp}) and $\mathcal{C}^e = (\mathcal{C}_i, \mathcal{C}_j)$ indicates $\mathcal{C}_i$ (e.g. \emph{chair}) is the parent concept of $\mathcal{C}_j$ (e.g. \emph{armchair}). Our DSSPN thus is constructed with the whole semantic neuron graph $\mathcal{G}^n= <\mathcal{N}^v, \mathcal{N}^e>$ with $M$ neurons in total. Each semantic neuron $n_i \in \mathcal{N}^v$ corresponds to one parent concept (e.g \emph{chair}) that has at least two child concepts within $\mathcal{C}^v$ and $\mathcal{N}^e$ corresponds to $\mathcal{C}^e$. As shown in Figure~\ref{fig:dynamic}, each neuron $n_i$ first employs one dense semantic-enhanced block to generate fine-grained features $\mathbf{h}_i$ using inherited features from its ancestors . The prediction layer $L_i$ with $1\times1$ convolutional filters takes $\mathbf{h}_i$ as input and produces $T_i$ prediction maps to distinguish between its $T_i$ child concepts $\{C_j\}, <\mathcal{C}^v_i, \mathcal{C}^v_j> \in \mathcal{C}^e$. Thus, each neuron is only responsible for recognizing a small set of confusing concepts by producting a distinct number of pixel-wise predictions. This hierarchical semantic propagation scheme significantly improves the model capacity for a large-scale concept vocabulary in the spirit of curriculum learning.

\subsection{Dense Semantic-enhanced Block}

Inspired from the successful practice of dense connectivity~\cite{huang2016densely} for image classification, we design a tree-structured dense semantic-enhanced block for improving information flow from the highly abstracted concepts to fine-grained concepts , following the inheritance path. Figure~\ref{fig:dynamic} illustrates the scheme of dense semantic-enhanced block for the \emph{desk} concept with the inheritance path: \emph{entity}-\emph{furniture-things}-\emph{table}-\emph{desk}. Let $\Omega_i$ denote the ancestor indexes of the concept $\mathcal{C}_i$ and $x$ as the convolutional features $\mathbf{x}$ from basic ConvNet. Consequently, each semantic neuron $n_i$ receives the feature maps $\{\mathbf{h}_{k}\}_{k\in \Omega_i}$ of all inherited preceding neurons $\{n_k\}_{k\in \Omega_i}$ starting from the root \emph{entity} concept as input:
\begin{equation}
\begin{split}
    \mathbf{h}_0 &= \mathcal{Q}(x),\\
    \mathbf{h}_i &= \mathcal{H}_{i}([ \{\mathbf{h}_{k}\}_{k\in\Omega_i}])
\end{split}
\end{equation}
where $\mathcal(Q)(\cdot)$ indicates the transition layer from basic convolutional features $x$ to features $\mathbf{h}_0$ of the root neuron. $[\{\mathbf{h}_{k}\}_{k\in\Omega_i}]$ refers to the concatenation of the feature maps produced in ancestral neurons $\{n_k\}_{k\in \Omega_i}$. $\mathcal{H}_i$ indicates the non-linear transformation function, composed of operations: rectified linear units (ReLU)~\cite{glorot2011deep} and Convolution (Conv). $\mathbf{h}_i$ is the resulting $m$ hidden feature maps of the neuron $n_i$. 

Each neuron $n_i$ thus has $m_0 +m\times(d - 1)$ input
feature maps, where $m_0 = 256$ is the channel number of hidden features $h_0$ of the root neuron after transitioning from $x$ and $d_i$ is the depth of concept $\mathcal{C}_i$ in the semantic concept hierarchy. $m$ is set as $48$ which is sufficient from our experiments in Section~\ref{sec:exp}. This information suppression with a relatively small number of hidden features can be regarded as retaining only details that are enough to distinguish between a small set of child concepts. 

Different from traditional semantic segmentation that learns one final prediction layer with a large number of feature maps to directly recognize all concepts, our DSSPN decomposes the pixel-wise predictions into a set of easier sub-tasks, which only needs a small feature map size for each sub-task and also improves the feature discriminative capability. An important difference between DSSPN and DenseNet~\cite{huang2016densely} is that DSSPN dynamically specifies different feature concatenation routes and depths for each concept following the concept knowledge graph. 

We design $\mathcal{Q}$ as as a Atrous Spatial Pyramid Pooling (ASSP) module ~\cite{chen2016deeplab} with three branches of $3\times3$ convolution layers and three rates as {6, 12, 18}, respectively. The output feature size of $\mathcal{Q}$ is $256$. The input feature size for $\mathcal{H}_i$ depends on the concept depth $d$, that is, the degree of fine-grained specification. In our case, the maximal depth is $5$, which effectively constrains the memory footprint growth. To improve computation efficiency, $\mathcal{H}_i$ first employs a bottleneck layer with a $1\times1$ convolution and a $3\times3$ convolution layer to reduce the number of input feature-maps, i.e., to the ReLU-Conv($1\times1$)-ReLU-Conv($3\times3$). Each semantic neuron and transition layer are followed by a ReLu function. The output feature size of $1\times1$ conv. layer is set as $4m$. 

\subsection{Dynamic-structured Semantic Propagation}
\label{sec:dynamic}
During training, our DSSPN performs the dynamic-structured semantic propagation, as illustrated in Figure~\ref{fig:universal}. Given a set of groundtruth concepts $\{\bar{\mathcal{C}}_t\}$ appearing in each image $I$, the activated semantic neuron structure $\bar{\mathcal{G}}^n = <\bar{\mathcal{N}}^v, \bar{\mathcal{N}}^e>$ can be obtained by only awakening semantic neurons $\bar{\mathcal{N}}^v = \{n_i\}, \mathcal{C}_i \in \beta(\{\bar{\mathcal{C}}_t\})$ and their edges that can derive the target labels. $\beta(\cdot)$ indicates the ancestor set of all groundtruth concepts. For example, for the second image in Figure~\ref{fig:universal}, only the neurons for the concepts \emph{entity}, \emph{structure-stuff}, \emph{plant-stuff} are activated in order to hierarchically segmenting out the targets \emph{tree}, \emph{grass}, \emph{plant-other}, \emph{fence} and \emph{animal-things}. Note that the neuron of \emph{animal-things} is deactivated since the image is only annotated with a coarse \emph{animal} class instead of more precise $dog$. Formally, the dynamic-structured neural computation graph can be constituted by recurrently propagating hidden features along the activate structure as:
\begin{equation}
	\begin{split}
	\centering
    \mathbf{h}_i &= \mathcal{H}_{i}([\mathbf{x}, \{\mathbf{h}_{k}\}_{k\in\Omega_i}]),\\
    P_i &= L_i(\mathbf{h}_i),\\
    \mathbf{h}_j &= \mathcal{H}_{j}([\mathbf{x}, \{\mathbf{h}_{k}\}_{k\in\Omega_i}, \mathbf{h_i}]), (n_i, n_j) \in \bar{\mathcal{N}}^e,
    \end{split}
\end{equation}
where the output hidden features $\mathbf{h}_i$ are only propagated to the activated child neurons in $\bar{\mathcal{N}}^v$ for each training image. Starting from the root neuron, our DSSPN recurrently traverses the whole activated semantic neuron sub-graph for hierarchical pixel-wise prediction. It thus leads to the dynamic-structured neural module back-propagation for each image. 

For training each neuron, we use the pixel-wise binary cross-entropy loss to supervise the dense prediction of each child concept, which focuses more on recognizing each child concept instead of learning any competition between them. This good characteristic leads a better flexibility for adding and pruning child concepts of each parent neuron, especially for joint training multiple datasets and extending the semantic concept hierarchy.

During testing phase, we use the hierarchical pixel-wise prediction over the semantic neuron graph. Starting from the root neuron, each neuron predicts the per-pixel predictions for classifying its child neurons and then only activates child ones with available predictions for further parsing regions.

\subsection{Universal Semantic Segmentation}

As shown in Figure~\ref{fig:universal}, our DSSPN can be naturally used to train a universal semantic segmentation for combining diverse segmentation dataset. The distinct concept sets from different dataset can be simply projected into a unified knowledge graph, and each image is then trained using the same strategy described in Section~\ref{sec:dynamic}. However, different datasets may be annotated with diverse granularities. For example,  the \emph{road} region on Cityscape dataset is further annotated into several fine-grained concepts on Mapillary dataset, e.g. \emph{curb}, \emph{crosswalk}, \emph{curb cut} and \emph{lane}. In order to alleviate the label discrepancy issues and stabilize the parameter optimization during joint training, we propose a concept-masking scheme. 

For training each image from the dataset $D_t$, we mask out the undefined concepts that share the same parent with defined concepts in $D_t$ during training. As a toy example, to train the third image in Figure 3, the \emph{way} neuron only outputs the pixel-wise predictions for \emph{road} and \emph{sidewalk} and ignores the predictions for undefined concepts in Cityscape, e.g. \emph{lane}. That way would thus improve the labeling consistency during joint training. 

Another merit of our DSSPN is the ability of updating and extending the model capacity in an online way. Benefiting the usage of dynamic-structured propagation scheme and joint training strategies, we can dynamically add and prune semantic neurons and concept labels for different purposes (e.g. adding more dataset) while keeping the previously learned parameters.

\subsection{Dynamic Batching Optimization}

Instead of using single instance training in most of tree-structured~\cite{socher2011parsing} and graph-structured~\cite{liang2016semantic,liang2017deep} networks, our DSSPN uses dynamic graph batching strategy to make good use of efficient data-parallel algorithms and hardware, inspired by very recent attempts~\cite{looks2017deep,bowman2016fast,neubig2017fly}.  We implement our DSSPN on PyTorch, which is one of dynamic neural network toolkit that offers more flexibility for coping with data of varying structures, compared to those that operate on statically declared computations (e.g.,
TensorFlow). Note that the neurons for high-level concepts (e.g. \emph{animal-things}) are executed more often than those for fine-grained concepts (e.g. \emph{dog}). For each batch, our DSSPN automatically batches those semantic neurons that are shared over all images for parallelism and then forwards execution to rest few isolated neurons following the activated neuron graph. DSSPN can thus speedups dynamically declared computation graphs for all images within one batch since most of shared semantic neurons are in place in the first few depths.

\subsection{Memory-Efficient Implementation}

Despite of the large whole semantic neruon graph of DSSN, it only activates a relative small computation graph for each image during training, which effectively constraints the memory and computation consumption. Although each semantic neuron only produces $m$ feature maps (where $m$ is small– set as $48$), but uses all previous feature maps from its ancestors as input. This would cause the number of parameters to grow quadratically with semantic hierarchy depth, which could be solved by a proper memory allocation strategy. To further reduce the memory consumption, we share memory allocations of neurons for parent concepts in practice. It effectively reduces the memory cost for storing feature maps from quadratic to linear.

\section{Experiments}
We empirically demonstrate DSSPN’s effectiveness on four benchmark semantic segmentation datasets and compare with state-of-the-art architectures

\subsection{Training Details}
\label{sec:training}
\textbf{Network Training.} We implement our DSSPN on Pytorch, with 2 GTX TITAN X 12GB cards on a single server. We use the Imagenet-pretrained ResNet-101~\cite{he2016deep} networks as our basic ConvNet following the procedure of~\cite{chen2016deeplab} and employ \emph{output stride = 8}, and replace the 1000-way Imagenet classifier in the last layer with our DSSPN structure. The network parameters in each neuron are presented in Section~\ref{sec:framework}, and the padding is set to keep the feature resolutions in DSSPN. We fix the moving means and variations in batch normalization layers of Resnet-101 during finetuning. The sum of binary cross-entropy loss for each position is employed on each semantic neuron to train hierarchical dense prediction. The predictions are thus compared with the ground truth labels (subsampled by 8), and the unlabeled pixels are ignored. We optimize the objective function with respect to the weights at all layers by the standard SGD procedure.Inspired by~\cite{chen2016deeplab}, we use the ``poly" learning rate policy and set base learning rate to 0.003 for newly initialized DSSPN parameters and power to 0.9. We set the learning rate as 0.00003 for pretrained layers. For each dataset, we train 90 epochs for the good convergence. Momentum and weight decay are set to 0.9 and 0.0001 respectively. For data augmentation, we adopt random flipping, random cropping and random resize between 0.5 and 2 for all datasets. Due to the GPU memory limitation, the batch size is 6 for Cityscape and Mapillary dataset, and 4 for Cocostuff and ADE20k dataset due to their larger label numbers (e.g. 182). The input crop size is set as $513\times513$ for all datasets.

We first evaluate the proposed DSSPN on four challenging datasets: Coco-Stuff~\cite{caesar2016coco}, ADE20k~\cite{zhou2016semantic}, Cityscape~\cite{cordts2015cityscapes} and Mapillary dataset~\cite{neuhold2017mapillary}. Note that we use the same DSSPN structure for all dataset during training. During testing, we only perform hierarhical pixe-wise prediction by only selecting a semantic neuron sub-graph that can reach out the defined concepts on each dataset. The mean IoU metrics are used for all datasets. We then evaluate its effectiveness on the universal semantic segmentation task by training a single model using all datasets.

\begin{table}[!tp]\setlength{\tabcolsep}{2pt}
	\centering\footnotesize{\caption{ Comparison with existing semantic segmentation models (\%) on the ADE20K val set~\cite{zhou2016semantic}. PSPNet (101)+DA+AL~\cite{zhao2016pyramid} used other data augmentation scheme and auxiliary loss. ``Conditional Softmax (VGG)~\cite{redmon2016yolo9000}",  ``Word2Vec(VGG)~\cite{frome2013devise}" and ``Joint-Cosine (VGG)~\cite{zhao2017open}" indicate existing approaches that also attempted the hierarchical classification, obtained from~\cite{zhao2017open}.}\label{tab:ade}
	\begin{tabular}{cccccccccccccccccccccc}
		\toprule
		{Method} & mean IoU &  pixel acc. \\
		\midrule
		FCN~\cite{long2015fully} & 29.39 & 71.32 \\
        SegNet~\cite{badrinarayanan2015segnet} & 21.64 & 71.00 \\
        DilatedNet~\cite{yu2015multi} &32.31 & 73.55\\
        CascadeNet~\cite{zhou2016semantic} & 34.90 & 74.52\\
		ResNet-101, 2 conv~\cite{wu2016wider}  & 39.40 & 79.07  \\
		PSPNet (ResNet-101)+DA+AL~\cite{zhao2016pyramid} & 41.96 & {80.64}\\
		\hline
		Conditional Softmax (VGG)~\cite{redmon2016yolo9000} (Hierarchical) &  31.27  & 72.23\\
		Word2Vec(VGG)~\cite{frome2013devise} (Hierarchical) & 29.18 & 71.31\\
		Joint-Cosine (VGG)~\cite{zhao2017open}(Hierarchical) & 31.52 & 73.15\\
		\midrule
		DSSPN (VGG)-Softmax & 31.01 & 73.20 \\
		DSSPN (VGG) & 34.56 & 76.04 \\
		{DSSPN (ResNet-101)-Softmax} & {39.23} & {78.57}\\
		{DSSPN (ResNet-101)} & {42.03} & {80.81}\\
		{DSSPN (ResNet-101) finetune} & {42.17} & {80.23}\\
		\textbf{DSSPN (ResNet-101) Universal} & \textbf{43.68} & \textbf{81.13}\\
		\hline
				\vspace{-7mm}
	\end{tabular}%
	}
\end{table}%

\textbf{Semantic Concept Hierarchy Construction.} We build the semantic concept hierarchy by combining labels from all four popular dataset. Starting from the label hierarchical tree of COCO-Stuff~\cite{caesar2016coco} that includes 182 concepts and 27 super-classes, we manually  merge concepts from the rest three dataset together by using WordTree. Note that we only add minimal number of intermediate super-classes during merging. It results in 359 concepts in the final concept hierarchical tree, as included in the supplementary materials. The maximal depth of resulting concept hierarchy is five. On average, six semantic neurons of DSSPN within each batch are activated for the images in COCO-Stuff, and 5 in ADE20k, 10 in Cityscape and 8 in Mapillary during training.

\subsection{Comparison with state-of-the-arts}

We directly apply the same hyper-parameters described in Section~\ref{sec:training} for clearly demonstrating the effectiveness of our dynamic-structure propagation network in general cases. Due to space limitation, we refer the readers to their dataset papers~\cite{zhou2016semantic,caesar2016coco,cordts2015cityscapes,neuhold2017mapillary} for different evaluation metrics.

\textbf{ADE20k dataset~\cite{zhou2016semantic}} consists of 20,210 images for training and 2,000 for validation. Images from both indoor and outdoor are annotated with 150 semantic concepts, including painting, lamp, sky, land, etc.
We first compare DSSPN with state-of-the-art methods that also use Resnet-101 as basic network in Table~\ref{tab:ade}. Our DSSPN performs better than the previous methods based on ResNet-101. Our DSSPN obtains 2.63\% higher mean IoU than the baseline model ``ResNet-101, 2 conv~\cite{wu2016wider}" that does multi-class recognition. We cannot fairly compare the state-of-the-arts~\cite{wu2016wider,zhao2016pyramid} since they used wider or deeper Imagenet pretrained networks. This clearly shows that incorporating dynamic-structured neurons can improve the model capacity for recognizing over a large concept vocabulary.

We further compare our DSSPN with prior works that also tried the hierarchical classification~\cite{redmon2016yolo9000,frome2013devise,zhao2017open} based on pretrained VGG
net, as reported in~\cite{zhao2017open}. Benefiting from learning distinct features for differentiating the child concepts of each super-class, ``DSSPN (VGG)-Softmax" that also uses Softmax loss on each semantic neuron significantly outperforms~\cite{redmon2016yolo9000,frome2013devise,zhao2017open} that simply utilized identical features for super-class categorizations at different levels.

\begin{table}[!tp]\setlength{\tabcolsep}{2pt}
	\centering\footnotesize{\caption{ Comparison on Coco-Stuff testing set (\%). All previous results are collected from~\cite{caesar2016coco}}\label{tab:coco}
	\begin{tabular}{cccccccccccccccccccccc}
		\toprule
		{Method} & Class-average acc. &  acc. & mean IoU \\
		\midrule
        FCN~\cite{long2015fully} & 38.5 & 60.4 & 27.2\\
        DeepLabv2 (ResNet-101)~\cite{chen2016deeplab} & 45.5 & 65.1 & 34.4\\
		DAG-RNN + CRF~\cite{shuai2017scene} & 42.8 & 63.0 & 31.2 \\
		OHE + DC + FCN~\cite{hu2017labelbank} & 45.8 & 66.6 & 34.3\\
		\midrule
		{DSSPN (ResNet-101)} & {47.0} & {68.5} & {36.2}\\
		{DSSPN (ResNet-101) finetune} & {48.1} & {69.4} & {37.3} \\
		\textbf{DSSPN (ResNet-101) Universal} & \textbf{50.3} & \textbf{70.7} & \textbf{38.9}\\
		\hline
				\vspace{-5mm}
	\end{tabular}%
	}
\end{table}%

\textbf{Coco-Stuff dataset~\cite{caesar2016coco}} contains 10,000 complex images from COCO with dense annotations of 91 thing and 91 stuff classes, including 9,000 for training and 1,000 for testing. We compare DSSPN with the state-of-the-art methods in Table~\ref{tab:coco}. We can observer DSSPN significantly outperforms existing methods. It further demonstrates that modeling semantic label hierarchy into network feature learning benefits for recognizing over a large vocabulary (e.g. 182) that can be hierarchically grouped into diverse super-classes . 

\begin{table}[!tp]\setlength{\tabcolsep}{2pt}
	\centering\footnotesize{\caption{ Comparison on Cityscapes testing set. }\label{tab:city}
	\begin{tabular}{cccccccccccccccccccccc}
		\toprule
		{Method} & IoU cla. &  iIoU cla. & IoU cat. & iIoU cat. \\
		\midrule
        FCN~\cite{long2015fully} & 65.3 & 41.7 & 85.7 & 70.1\\
        LRR~\cite{ghiasi2016laplacian} & 69.7 & 48.0 & 88.2 & 74.7\\
        DeepLabv2 (ResNet-101)~\cite{chen2016deeplab} & 70.4 & 42.6 & 86.4 & 67.7\\
		Piecewise~\cite{lin2016efficient}  & 71.6 & 51.7 & 87.3 & 74.1 \\
		\midrule
		{DSSPN (ResNet-101)} & {74.0} & {53.5} & {88.5} & {76.1}\\
		{DSSPN (ResNet-101) finetune} & {74.6} & {53.9} & {89.1} & {77.0}\\
		\textbf{DSSPN (ResNet-101) Universal} & \textbf{76.6} & \textbf{56.2} & \textbf{89.6} & \textbf{77.8}\\
		\hline
				\vspace{-5mm}
	\end{tabular}%
	}
\end{table}%
\textbf{Cityscape dataset~\cite{cordts2015cityscapes}} contains 5,000 urban scene images collected from 50 cities, which are splited into 2,975, 500, and 1,525 for training, validation and testing. The pixel-wise annotations of 19 concepts (e.g. \emph{road}, \emph{fence}) are provided. We reports results on Cityscape test set in Table~\ref{tab:city}. Our DSSPN is also based on ResNet101 using single scale inputs for testing and does not employ post-processing like CRF as in our fair baseline ``DeepLabv2 (ResNet-101)~\cite{chen2016deeplab}". Compared to our fair baseline, ``DSSPN (ResNet-101)" brings significant improvement, i.e. 3.6\% in IoU class. Note that we cannot fairly compare with recent best performances~\cite{zhao2016pyramid,wu2016wider} on Cityscape benchmark since they often combined results from several scaled inputs or used different base models.

\begin{table}[!tp]\setlength{\tabcolsep}{2pt}
	\centering\footnotesize{\caption{ Comparison on Mapillary validation set (\%). The results of~\cite{wu2016wider,zhao2016pyramid} are reported in~\cite{neuhold2017mapillary}. }\label{tab:mapillary}
	\begin{tabular}{cccccccccccccccccccccc}
		\toprule
		{Method} &  mean IoU \\
		\midrule
        Wider Network~\cite{wu2016wider}  & 41.12\\
        PSPNet~\cite{zhao2016pyramid} & 49.76\\
        Baseline ResNet-101           & 37.58\\
		\midrule
		{DSSPN (ResNet-101)} & {42.39} \\
		{DSSPN (ResNet-101) finetune} & {42.57} \\
		\textbf{DSSPN (ResNet-101) Universal} & \textbf{45.01} \\
		\hline
				\vspace{-5mm}
	\end{tabular}%
	}
\end{table}%

\begin{figure}[!tp]
		\begin{center}
			\includegraphics[scale=0.45]{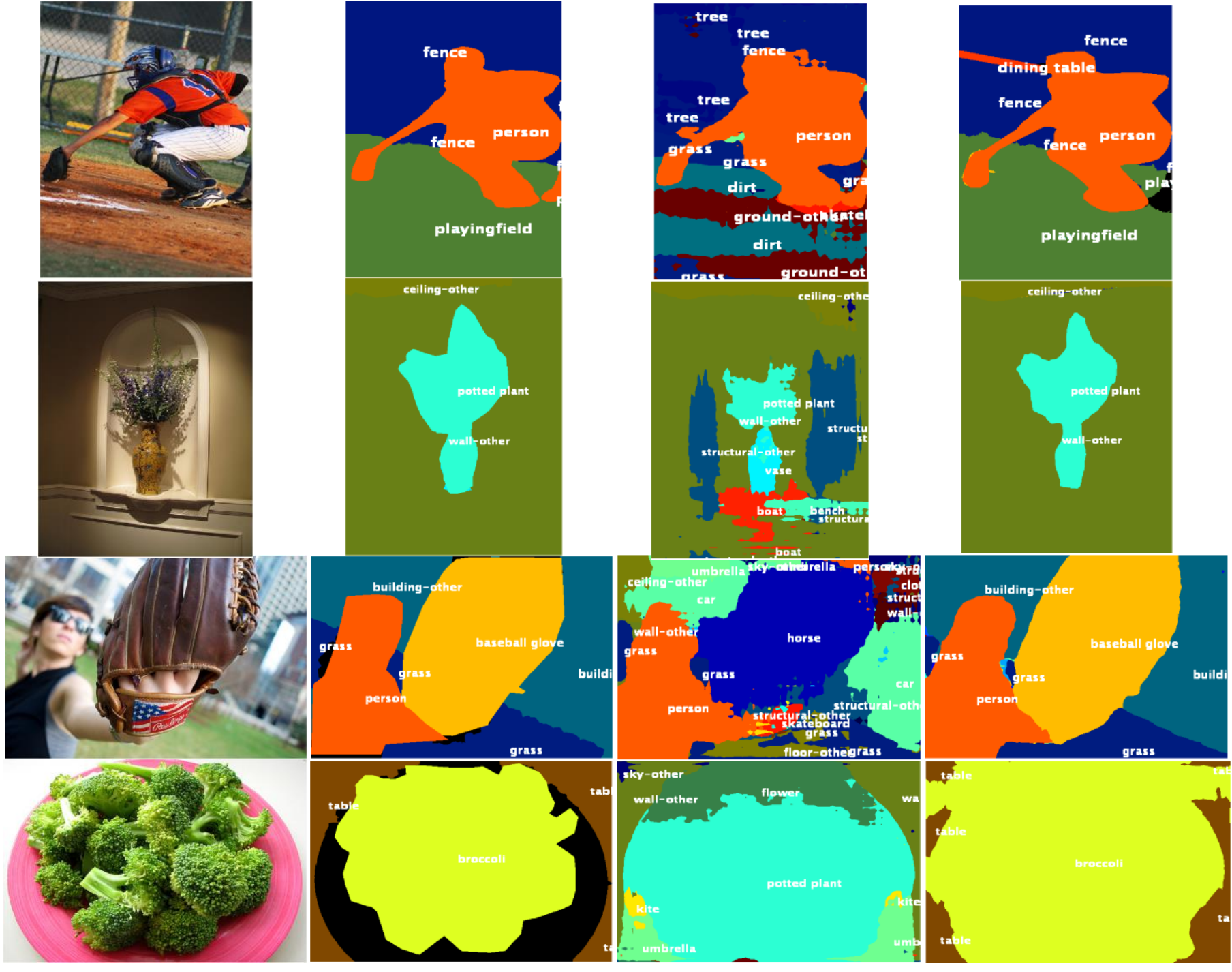}\vspace{-1mm}
			\caption{Visual comparison on Coco-Stuff. For each image, we sequentially show its groudtruth labeling, prediction of ``DSSPN (ResNet-101)", prediction of ``DSSPN (ResNet-101) Universal".}
			\label{fig:coco}
		\end{center}
		\vspace{-2mm}
	\end{figure}
	
\textbf{Mapillary dataset~\cite{neuhold2017mapillary}} includes 20,000 street-level images annotated into 66 object categories (e.g. rider, streetlight, traffic sign back), in which 18,000 are used for training and 2,000 for validation. We report the result comparisons in Table~\ref{tab:mapillary}. We mainly compare our DSSPN with the baseline ``ResNet-101" instead of previous methods~\cite{wu2016wider,zhao2016pyramid} since they used different basic networks. The large improvement by our DSSPN can be again observed, i.e. 4.81\% on mean IoU.

\subsection{Universal Semantic Segmentation Model}

Another interesting advantage of our DSSPN is its ability of training all dataset annotations within a unified segmentation model as ``DSSPN (ResNet-101) Universal". To train a unified model, we combine all training samples from four datasets and select images from the same dataset to construct one batch at each step. Since images on each dataset are collected from different scenarios and domains, we first train a unified model using all images for 60 epoch, and then decrease the learning rate by 1/10 to further finetune models for 20 epochs on each dataset. We reports results on each dataset in Table~\ref{tab:ade},~\ref{tab:coco},~\ref{tab:city},~\ref{tab:mapillary}, respectively. 

The commonly used strategy for utilizing other dataset annotations~\cite{chen2016deeplab,zhao2016pyramid} is to remove the final classification layer and retrain the newly initialized layers for new label sets due to the label discrepancy. Following such strategy, we also report the results of ``DSSPN (ResNet-101) finetune" that first train models on one dataset, and then finetune on the new dataset by retaining only network parameters of basic ConvNet. Since such strategy cannot support training on more than one dataset at once, we thus use the COCO-Stuff and ADE20k pair due to their similar image resources, and Cityscape and Mapillary training pair.

By comparing results of our ``DSSPN (ResNet-101) Universal" with ``DSSPN (ResNet-101) finetune" in all Tables, it can be demonstrated that jointly training all semantic neurons with diverse image annotations and resources can significantly improve the model generalization capability and thus leads to superior performance on each dataset. Figure~\ref{tab:coco} shows their quantitative comparisons on Coco-Stuff dataset. ``DSSPN (ResNet-101) Universal" yields more accurate concept recognition, especifically for some rare labels (e.g. \emph{baseball glove}, \emph{broccoli}).

\subsection{Discussion}

Table~\ref{tab:ablation} shows the ablation studies of our DSSPN to validate the effectiveness of its main components.

\begin{table}[!tp]\setlength{\tabcolsep}{2pt}
	\centering\footnotesize{\caption{Ablation studies on the ADE20K val set~\cite{zhou2016semantic}. }\label{tab:ablation}
	\begin{tabular}{cccccccccccccccccccccc}
		\toprule
		{Method} & mean IoU &  pixel acc.\\
		\midrule
		DSSPN fixed & 42.15 & 81.02 \\
		DSSPN w/o dense block & 38.41 & 76.59\\
		DSSPN w/ dense block, summation & 41.74 & 80.27\\
		DSSPN dense block (32) & 41.12 & 79.85\\
		DSSPN dense block (64) & 42.02 & 80.76\\
		DSSPN (synset in~\cite{zhou2016semantic}) & 41.96 & 81.21\\
		\midrule
		{DSSPN} & {42.03} & {80.81}\\
		\hline
				\vspace{-5mm}
	\end{tabular}%
	}
\end{table}%

\textbf{Dynamic-structure propagation.} The semantic propagation network can also be trained using a fixed structure where images must be passed through all semantic neurons needed for each dataset during training. No noticeable improvement can be seen by comparing ``DSSPN fixed" with our dynamic version while  ``DSSPN fixed" needs more computation  and memory cost. Since we use the hierarchical prediction scheme during testing, our DSSPN can be efficiently learned by only focusing on differentiating confusing concepts at each level of semantic neuron graph.

\textbf{Dense semantic-enhanced block.} An alternative basic block for each semantic neuron can be the directly feature propagation without dense connection, as ``DSSPN w/o dense block". Our experiment shows that it would sacrifice the performance but reduce the parameter numbers. We further demonstrate the feature concatenation used in dense block outperforms the feature summation version by comparing ``DSSPN w/ dense block, summation" with ours. For the hyper-parameter $m$ of hidden feature map size, we also evaluate the results of setting $m$ as 32 and 64. It can be seen that using moderately small feature map size ($m=48$) is sufficient for capturing key feature characteristics, which are used in all experiments on different datasets.

\textbf{Binary cross-entropy vs Softmax.} In Table~\ref{tab:ade}, we also show that using per-pixel sigmoid with binary cross-entropy loss (our model) significantly outperforms the Softmax loss that is common practice in other hierarchical classification model~\cite{redmon2016yolo9000}. The similar conclusion has been shown in Mask R-CNN~\cite{he2017mask} for instance-level segmentation that predicts a binary mask for each class independently, without competition among classes. The class competition by Softmax loss also hinders the model's capability of learning a unified model using diverse label annotations, where only some parts of concepts belonging to one super-class are visible. 

\textbf{The affect of different concept hierarchies.} Another interesting point that may be raised is how different concept graphs influence the final performance. We thus try the synset provided in original ADE20k dataset~\cite{zhou2016semantic} as the whole concept hierarchy and the results are reported as ``(synset in~\cite{zhou2016semantic})". We can observe that there is only slight performance changes by using the original synset tree in~\cite{zhou2016semantic} that includes more hypernyms for grouping the object affordance. 

\textbf{Model and computation complexity.}
\begin{table}[!tp]\setlength{\tabcolsep}{2pt}
	\centering\footnotesize{\caption{We show number of model parameters, average testing speed (img/sec) on the full image resolution, and mean IOU (\%) on Cityscape validation set. All results are evaluated using Pytorch under the same setting. }\label{tab:complex}
	\begin{tabular}{cccccccccccccccccccccc}
		\toprule
		{Model} & Params & Test-speed & mean IoU\\
		\midrule
		Deeplabv2 (ResNet-101)~\cite{chen2016deeplab} & 176.6M & 1.78 &71.0\\
		DSSPN (ResNet-50) & 141.0M & 2.26 & 73.2\\
		DSSPN (ResNet-101) & 217.3M & 1.45 & {75.5}\\
		\hline
				\vspace{-7mm}
	\end{tabular}%
	}
\end{table}%

In Table~\ref{tab:complex}, we report experiments with the baseline model ``Deeplabv2 (ResNet-101)" and our DSSPN variants on Cityscape validatation set for comparing their model sizes and time efficiency. Both our DSSPN variants using ResNet-50 and ResNet-101 yield much better performance than the baseline model. Moreover, ``DSSPN (Resnet-50)" reduces both computation consumption and model size compared to the baseline model. It should be noted that although DSSPN has more parameters by taking into account all semantic neurons within the graph, it only activates a small sub-set of neurons for each image during training and testing, benefiting from the dynamic-structured semantic propagation scheme.

\subsection{Conclusion and Future work}

In this paper, we proposed a novel dynamic-structured semantic propagation network for the general semantic segmentation tasks. Our DSSPN explicitly constructs a semantic neuron graph network by incorporating the semantic concept hierarchy. A dynamic-structured network optimization is performed to dynamically activate semantic neuron sub-graphs for each image during training. Extensive experiments on four public benchmarks demonstrate the superiority of our DSSPN. We further show our DSSPN can be naturally used to train a unified segmentation model over all available segmentation annotations, leading to its better generalization capability. In future, we plan to generalize DSSPN to other vision tasks and investigate how to embed more complex semantic relationships naturally into the network design. The proposed DSSPN is general enough to handle more complex semantic concept graphs that contain categories with multiple ancestors in the hierarchy. In that case,
each semantic neuron can simply combine features passed
from multiple ancestor neurons via summation, and then
performs the dynamic pixel-wise prediction.

\label{sec:exp}
{\small
\bibliographystyle{ieee}
\bibliography{egbib}

\begin{thebibliography}{10}\itemsep=-1pt

\bibitem{badrinarayanan2015segnet}
V.~Badrinarayanan, A.~Kendall, and R.~Cipolla.
\newblock Segnet: A deep convolutional encoder-decoder architecture for image
  segmentation.
\newblock In {\em CVPR}, 2015.

\bibitem{bengio2009curriculum}
Y.~Bengio, J.~Louradour, R.~Collobert, and J.~Weston.
\newblock Curriculum learning.
\newblock In {\em ICML}, pages 41--48, 2009.

\bibitem{bowman2016fast}
S.~R. Bowman, J.~Gauthier, A.~Rastogi, R.~Gupta, C.~D. Manning, and C.~Potts.
\newblock A fast unified model for parsing and sentence understanding.
\newblock {\em arXiv preprint arXiv:1603.06021}, 2016.

\bibitem{caesar2016coco}
H.~Caesar, J.~Uijlings, and V.~Ferrari.
\newblock Coco-stuff: Thing and stuff classes in context.
\newblock {\em arXiv preprint arXiv:1612.03716}, 2016.

\bibitem{chen2016deeplab}
L.-C. Chen, G.~Papandreou, I.~Kokkinos, K.~Murphy, and A.~L. Yuille.
\newblock Deeplab: Semantic image segmentation with deep convolutional nets,
  atrous convolution, and fully connected crfs.
\newblock {\em arXiv preprint arXiv:1606.00915}, 2016.

\bibitem{cordts2015cityscapes}
M.~Cordts, M.~Omran, S.~Ramos, T.~Scharw{\"a}chter, M.~Enzweiler, R.~Benenson,
  U.~Franke, S.~Roth, and B.~Schiele.
\newblock The cityscapes dataset.
\newblock In {\em CVPR Workshop}, volume~1, page~3, 2015.

\bibitem{deng2014large}
J.~Deng, N.~Ding, Y.~Jia, A.~Frome, K.~Murphy, S.~Bengio, Y.~Li, H.~Neven, and
  H.~Adam.
\newblock Large-scale object classification using label relation graphs.
\newblock In {\em ECCV}, pages 48--64, 2014.

\bibitem{deng2012hedging}
J.~Deng, J.~Krause, A.~C. Berg, and L.~Fei-Fei.
\newblock Hedging your bets: Optimizing accuracy-specificity trade-offs in
  large scale visual recognition.
\newblock In {\em CVPR}, pages 3450--3457, 2012.

\bibitem{everingham2010pascal}
M.~Everingham, L.~Van~Gool, C.~K. Williams, J.~Winn, and A.~Zisserman.
\newblock The pascal visual object classes (voc) challenge.
\newblock {\em IJCV}, 88(2):303--338, 2010.

\bibitem{frome2013devise}
A.~Frome, G.~S. Corrado, J.~Shlens, S.~Bengio, J.~Dean, T.~Mikolov, et~al.
\newblock Devise: A deep visual-semantic embedding model.
\newblock In {\em NIPS}, pages 2121--2129, 2013.

\bibitem{ghiasi2016laplacian}
G.~Ghiasi and C.~C. Fowlkes.
\newblock Laplacian pyramid reconstruction and refinement for semantic
  segmentation.
\newblock In {\em ECCV}, pages 519--534, 2016.

\bibitem{glorot2011deep}
X.~Glorot, A.~Bordes, and Y.~Bengio.
\newblock Deep sparse rectifier neural networks.
\newblock In {\em Proceedings of the Fourteenth International Conference on
  Artificial Intelligence and Statistics}, pages 315--323, 2011.

\bibitem{he2017mask}
K.~He, G.~Gkioxari, P.~Doll{\'a}r, and R.~Girshick.
\newblock Mask r-cnn.
\newblock {\em ICCV}, 2017.

\bibitem{he2016deep}
K.~He, X.~Zhang, S.~Ren, and J.~Sun.
\newblock Deep residual learning for image recognition.
\newblock In {\em CVPR}, pages 770--778, 2016.

\bibitem{hu2017labelbank}
H.~Hu, Z.~Deng, G.-T. Zhou, F.~Sha, and G.~Mori.
\newblock Labelbank: Revisiting global perspectives for semantic segmentation.
\newblock {\em arXiv preprint arXiv:1703.09891}, 2017.

\bibitem{huang2016densely}
G.~Huang, Z.~Liu, K.~Q. Weinberger, and L.~van~der Maaten.
\newblock Densely connected convolutional networks.
\newblock In {\em CVPR}, 2017.

\bibitem{kaiser2017one}
L.~Kaiser, A.~N. Gomez, N.~Shazeer, A.~Vaswani, N.~Parmar, L.~Jones, and
  J.~Uszkoreit.
\newblock One model to learn them all.
\newblock {\em arXiv preprint arXiv:1706.05137}, 2017.

\bibitem{kokkinos2016ubernet}
I.~Kokkinos.
\newblock Ubernet: Training auniversal'convolutional neural network for low-,
  mid-, and high-level vision using diverse datasets and limited memory.
\newblock In {\em CVPR}, 2017.

\bibitem{liang2017deep}
X.~Liang, L.~Lee, and E.~P. Xing.
\newblock Deep variation-structured reinforcement learning for visual
  relationship and attribute detection.
\newblock In {\em CVPR}, 2017.

\bibitem{liang2017interpretable}
X.~Liang, L.~Lin, X.~Shen, J.~Feng, S.~Yan, and E.~P. Xing.
\newblock Interpretable structure-evolving lstm.
\newblock In {\em CVPR}, 2017.

\bibitem{liang2016semantic}
X.~Liang, X.~Shen, J.~Feng, L.~Lin, and S.~Yan.
\newblock Semantic object parsing with graph lstm.
\newblock In {\em ECCV}, 2016.

\bibitem{lin2016refinenet}
G.~Lin, A.~Milan, C.~Shen, and I.~Reid.
\newblock Refinenet: Multi-path refinement networks with identity mappings for
  high-resolution semantic segmentation.
\newblock In {\em CVPR}, 2017.

\bibitem{lin2016efficient}
G.~Lin, C.~Shen, A.~van~den Hengel, and I.~Reid.
\newblock Efficient piecewise training of deep structured models for semantic
  segmentation.
\newblock In {\em CVPR}, pages 3194--3203, 2016.

\bibitem{long2015fully}
J.~Long, E.~Shelhamer, and T.~Darrell.
\newblock Fully convolutional networks for semantic segmentation.
\newblock In {\em CVPR}, pages 3431--3440, 2015.

\bibitem{looks2017deep}
M.~Looks, M.~Herreshoff, D.~Hutchins, and P.~Norvig.
\newblock Deep learning with dynamic computation graphs.
\newblock {\em arXiv preprint arXiv:1702.02181}, 2017.

\bibitem{neubig2017fly}
G.~Neubig, Y.~Goldberg, and C.~Dyer.
\newblock On-the-fly operation batching in dynamic computation graphs.
\newblock In {\em NIPS}, 2017.

\bibitem{neuhold2017mapillary}
G.~Neuhold, T.~Ollmann, S.~Rota~Bulo, and P.~Kontschieder.
\newblock The mapillary vistas dataset for semantic understanding of street
  scenes.
\newblock In {\em CVPR}, pages 4990--4999, 2017.

\bibitem{niepert2016learning}
M.~Niepert, M.~Ahmed, and K.~Kutzkov.
\newblock Learning convolutional neural networks for graphs.
\newblock In {\em ICML}, pages 2014--2023, 2016.

\bibitem{noh2015learning}
H.~Noh, S.~Hong, and B.~Han.
\newblock Learning deconvolution network for semantic segmentation.
\newblock In {\em ICCV}, pages 1520--1528, 2015.

\bibitem{ordonez2013large}
V.~Ordonez, J.~Deng, Y.~Choi, A.~C. Berg, and T.~L. Berg.
\newblock From large scale image categorization to entry-level categories.
\newblock In {\em ICCV}, pages 2768--2775, 2013.

\bibitem{redmon2016yolo9000}
J.~Redmon and A.~Farhadi.
\newblock Yolo9000: better, faster, stronger.
\newblock In {\em CVPR}, 2017.

\bibitem{shi2017deep}
X.~Shi, Z.~Gao, L.~Lausen, H.~Wang, D.-Y. Yeung, W.-k. Wong, and W.-c. Woo.
\newblock Deep learning for precipitation nowcasting: A benchmark and a new
  model.
\newblock In {\em NIPS}, 2017.

\bibitem{shuai2017scene}
B.~Shuai, Z.~Zuo, B.~Wang, and G.~Wang.
\newblock Scene segmentation with dag-recurrent neural networks.
\newblock {\em TPAMI}, 2017.

\bibitem{socher2011parsing}
R.~Socher, C.~C. Lin, C.~Manning, and A.~Y. Ng.
\newblock Parsing natural scenes and natural language with recursive neural
  networks.
\newblock In {\em ICML}, pages 129--136, 2011.

\bibitem{tishby2015deep}
N.~Tishby and N.~Zaslavsky.
\newblock Deep learning and the information bottleneck principle.
\newblock In {\em Information Theory Workshop (ITW)}, pages 1--5, 2015.

\bibitem{wang2016cnn}
J.~Wang, Y.~Yang, J.~Mao, Z.~Huang, C.~Huang, and W.~Xu.
\newblock Cnn-rnn: A unified framework for multi-label image classification.
\newblock In {\em CVPR}, pages 2285--2294, 2016.

\bibitem{wu2016wider}
Z.~Wu, C.~Shen, and A.~v.~d. Hengel.
\newblock Wider or deeper: Revisiting the resnet model for visual recognition.
\newblock {\em arXiv preprint arXiv:1611.10080}, 2016.

\bibitem{yu2015multi}
F.~Yu and V.~Koltun.
\newblock Multi-scale context aggregation by dilated convolutions.
\newblock {\em arXiv preprint arXiv:1511.07122}, 2015.

\bibitem{zhao2017open}
H.~Zhao, X.~Puig, B.~Zhou, S.~Fidler, and A.~Torralba.
\newblock Open vocabulary scene parsing.
\newblock In {\em ICCV}, 2017.

\bibitem{zhao2016pyramid}
H.~Zhao, J.~Shi, X.~Qi, X.~Wang, and J.~Jia.
\newblock Pyramid scene parsing network.
\newblock In {\em CVPR}, 2017.

\bibitem{zhou2016semantic}
B.~Zhou, H.~Zhao, X.~Puig, S.~Fidler, A.~Barriuso, and A.~Torralba.
\newblock Semantic understanding of scenes through the ade20k dataset.
\newblock {\em arXiv preprint arXiv:1608.05442}, 2016.

\end{thebibliography}
}

\end{document}